\documentclass[preprint,12pt]{elsarticle}
\usepackage{graphicx}
\usepackage{enumerate}
\usepackage{booktabs}
\usepackage{multirow}
\usepackage{multicol} 
\usepackage{makecell}
\usepackage[table]{xcolor}
\usepackage{amsmath, amsfonts, amssymb, bm}
\usepackage{hyperref}
\usepackage{cleveref}
\usepackage{caption}
\usepackage{subcaption}
\usepackage{listings}
\usepackage[misc,geometry]{ifsym}
\usepackage[margin=3cm]{geometry}

\graphicspath{{img}}

\newcommand{\rev}[1]{{\color{black}#1}}

\begin{document}
	
	\begin{frontmatter}
		
		\title{TOC-UCO: a comprehensive repository of tabular ordinal classification datasets}
		
		
		\author[IMIBIC,DOCTORADO]{Rafael Ayllón-Gavilán\corref{cor1}} 
		\author[UCO]{David Guijo-Rubio}
		\author[UCO]{Antonio Manuel Gómez-Orellana}
		\author[UCO]{Francisco Bérchez-Moreno}
		\author[UCO]{Víctor Manuel Vargas-Yun}
		\author[UCO]{Pedro A. Gutiérrez}
		
		
		\cortext[cor1]{Corresponding author\\ E-mail address: \url{rafael.ayllon@imibic.org}}
		\address[IMIBIC]{Department of Clinical-Epidemiological Research in Primary Care, Instituto Maimónides de Investigación Biomédica de Córdoba (IMIBIC), Spain.}
		\address[DOCTORADO]{``Programa de doctorado en Computación Avanzada, Energía y Plasmas'', Universidad de Córdoba, Córdoba, Andalucía, Spain.}
		\address[UCO]{Departamento de Ciencia de la Computación e Inteligencia Artificial, Universidad de Córdoba, Spain.}
		
		
		\begin{abstract}
			An ordinal classification (OC) problem corresponds to a special type of classification characterised by the presence of a natural order relationship among the classes. This type of problem can be found in a number of real-world applications, motivating the design and development of many ordinal methodologies over the last years. However, it is important to highlight that the development of the OC field suffers from one main disadvantage: the lack of a comprehensive set of datasets on which novel approaches to the literature have to be benchmarked. In order to approach this objective, this manuscript from the University of Córdoba (UCO), which have previous experience on the OC field, provides the literature with a publicly available repository of tabular data for a robust validation of novel OC approaches, namely TOC-UCO (Tabular Ordinal Classification repository of the UCO). Specifically, this repository includes a set of $46$ tabular ordinal datasets, preprocessed under a common framework and ensured to have a reasonable number of patterns and an appropriate class distribution. We also provide the sources and preprocessing steps of each dataset, along with details on how to benchmark a novel approach using the TOC-UCO repository. For this, indices for $30$ different randomised train-test partitions are provided to facilitate the reproducibility of the experiments.
		\end{abstract}
		
		\begin{keyword}
			ordinal classification \sep ordinal regression \sep tabular data \sep benchmarking \sep data
		\end{keyword}
		
	\end{frontmatter}
	
	\section{Introduction}
	Ordinal classification (OC) covers those classification problems in which the output labels (classes) present an order relationship among them. For example, when classifying patients according to the stage of a particular disease, the labels can be \{\textit{Healthy, Mild, Moderate, Moderately Severe, Severe}\}. Note that the distance between those classes can be undefined and unequal, i.e. the distance or similarity between \textit{Healthy} and \textit{Mild} patients can be lower than the distance between \textit{Mild} and \textit{Moderate} ones. Commonly, OC problems are addressed using standard nominal approaches, which often reach non-optimal solutions. The field of OC is focused on developing specialised techniques able to exploit the ordinal information of the classes, with the aim of achieving more competitive solutions in terms of ordinal performance. This field is also known as ordinal regression.
	
	
	\rev{In the current OC literature, the standard set of problems utilised for benchmarking is the one presented in \cite{orreview}, where the authors performed an exhaustive search over the internet to collect publicly available ordinal data. This set of problems conforms the baseline for TOC-UCO, and will be referred to as ``Previous'' in the rest of the manuscript. For instance, some of the datasets of this Previous repository were collected from \cite{chu2007support}. The metadata of Previous is presented in \Cref{tab:orreview_data}, comprising a total of 29 unique databases, where 17 are originally OC datasets, and the remaining 12 are regression datasets whose output was discretised in either 5 or 10 bins. Some datasets from Previous have gained significant popularity in the field of OC and have been used for the evaluation and validation of several methodological contributions \cite{liu2024distributed,lei2024core}. However, it requires improvements in several aspects. In the first place, many of the datasets present a very high imbalance, that is often combined with a low number of patterns (e.g. the \textit{contact-lenses} or \textit{squash-unstored} datasets). In addition, the discretised regression problems are repeated but with a different number of classes, i.e. bins in the discretisation process, which can bias the results in favour of the methodologies that have a good performance on them. Furthermore, an equal-frequency discretisation was performed on the datasets in Previous, which results in a non natural class distribution, where all classes have the exact same number of patterns.
		
		Up to date, in spite of the popularity of the datasets in Previous, it is a detrimental but widespread practice to limit the experiments to only a subset of them when validating new methodologies. Moreover, the specific subset of datasets used often varies across different studies. This may imply several issues, such as \textit{cherry-picking}, i.e. selecting from a larger set of datasets only those where the proposed approach outperforms previous methods, or omitting important factors to be considered such as model complexity (e.g. using small datasets or those with a reduced number of classes), among other possibilities. We have observed that the validation of many recent methodologies still suffer from the aforementioned drawbacks, as an unified set of datasets is not used to benchmark novel approaches. 
		
		The problem introduced above may be due to the fact that the Previous repository was not formally presented, but was simply the collection of data used in the associated experimental study. With the current manuscript, we aim to present and popularise the new TOC-UCO repository as the standard benchmarking set of OC problems, so that it becomes established in the OC literature, alleviating the aforementioned issue.
	}
	
	\begin{table}
		\footnotesize
		\setlength{\tabcolsep}{4pt}
		\renewcommand{\arraystretch}{0.75}
		\caption{Characteristics of the ordinal classification datasets provided in \cite{orreview}. The class distribution of the discretised regression datasets is uniform, i.e. each class has approximately the same number of patterns. $N$ represents the number of patterns, $K$ the number of input features, and $Q$ the number of classes.}
		\label{tab:orreview_data}
		\centering
		\small
		\begin{tabular}{lrrrr}
			\toprule
			\multicolumn{5}{c}{Discretised regression datasets}\\
			\midrule
			Dataset	&	$N$	&	$K$	&	$Q$	&	Class distribution\\
			\midrule
			pyrim5 &	74	&	27	&	5	&	~15	 per class \\
			machine5 &	209	&	7	&	5	&	~42	 per class \\
			housing5 &	506	&	14	&	5	&	~101	 per class \\
			stock5 &	700	&	9	&	5	&	140	 per class \\
			abalone5 &	4177	&	11	&	5	&	~836	 per class \\
			bank5 &	8192	&	8	&	5	&	~1639	 per class \\
			bank5' &	8192	&	32	&	5	&	~1639	 per class \\
			computer5 &	8192	&	12	&	5	&	~1639	 per class \\
			computer5' &	8192	&	21	&	5	&	~1639	 per class \\
			cal.housing5 &	20640	&	8	&	5	&	4128	 per class \\
			census5 &	22784	&	8	&	5	&	~4557	 per class \\
			census5' &	22784	&	16	&	5	&	~4557	 per class \\
			pyrim10 &	74	&	27	&	10	&	~8	 per class \\
			machine10 &	209	&	7	&	10	&	~21	 per class \\
			housing10 &	506	&	14	&	10	&	~51	 per class \\
			stock10 &	700	&	9	&	10	&	70	 per class \\
			abalone10 &	4177	&	11	&	10	&	~418	 per class \\
			bank10 &	8192	&	8	&	10	&	~820	 per class \\
			bank10' &	8192	&	32	&	10	&	~820	 per class \\
			computer10 &	8192	&	12	&	10	&	~820	 per class \\
			computer10' &	8192	&	21	&	10	&	~820	 per class \\
			cal.housing10 &	20640	&	8	&	10	&	2064	 per class \\
			census10 &	22784	&	8	&	10	&	~2279	 per class \\
			census10' &	22784	&	16	&	10	&	~2279	 per class \\
			\midrule
			\multicolumn{5}{c}{Originally ordinal classification datasets}\\
			\midrule
			Dataset	&	$N$	&	$K$	&	$Q$	&	Class distribution \\
			\midrule    
			contact-lenses	&	24	&	6	&	3	&	(15 5 4) \\
			pasture		&	36	&	25	&	3	&	(12 12 12) \\
			squash-stored	&	52	&	51	&	3	&	(23 21 8) \\
			squash-unstored	&	52	&	52	&	3	&	(24 24 4) \\
			tae		&	151	&	54	&	3	&	(49 50 52) \\
			newthyroid		&	215	&	5	&	3	&	(30 150 35) \\
			balance-scale	&	625	&	4	&	3	&	(288 49 288) \\
			SWD		&	1000	&	10	&	4	&	(32 352 399 217) \\
			car		&	1728	&	21	&	4	&	(1210 384 69 65) \\
			bondrate		&	57	&	37	&	5	&	(6 33 12 5 1) \\
			toy		&	300	&	2	&	5	&	(35 87 79 68 31) \\
			eucalyptus		&	736	&	91	&	5	&	(180 107 130 214 105) \\
			LEV		&	1000	&	4	&	5	&	(93 280 403 197 27) \\
			automobile		&	205	&	71	&	6	&	(3 22 67 54 32 27) \\
			winequality-red	&	1599	&	11	&	6	&	(10 53 681 638 199 18) \\
			ESL		&	488	&	4	&	9	&	(2 12 38 100 116 135 62 19 4) \\
			ERA		&	1000	&	4	&	9	&	(92 142 181 172 158 118 88 31 18) \\
			\bottomrule
		\end{tabular}
	\end{table}
	
	The TOC-UCO repository presented in this work addresses the aforementioned gaps, aiming to provide a new improved and updated set of benchmark problems to the OC literature. In the following, we present the improvements introduced by this repository:
	\begin{itemize}
		\item It presents a higher number of datasets, from the previous 29, to 46 unique ordinal datasets.
		\item Every dataset presents an appropriate number of patterns in each of the classes. The less represented class over the whole repository has 16 patterns. In the Previous repository, there were classes represented by less than 4 patterns, including classes represented by just a single pattern (\Cref{tab:orreview_data}).
		\item The discretised regression problems are not duplicated with a different number of classes.
		\item The regression datasets are discretised following a clustering approach for establishing the thresholds separating each class, that results in a more natural distribution of the classes.
		\item It presents a high diversity in the number of classes: 4 datasets of 3 classes, 11 datasets of 4 classes, 10 datasets of 5 classes, 15 datasets from 6 to 9 classes, and 6 datasets with 10 classes.
	\end{itemize}
	
	\rev{In this way, the main contributions of this work can be listed as follows:
		\begin{itemize}
			\item To provide the ordinal classification literature with TOC-UCO, a new benchmark repository comprising a high number of datasets with diverse characteristics, preprocessed under an unified framework.
			\item To provide the analysis, description, and source of each dataset.
			\item To provide the Python code to load and utilise TOC-UCO.
			\item To provide baseline results of nominal and ordinal models in TOC-UCO.
			\item To provide 30 different fixed train-test partitions, enabling the reproducibility of experiments.
		\end{itemize}
	}
	
	\rev{The rest of this work is organised as follows: \Cref{sec:2_oc} formally introduces OC, together with the main challenges that arise when dealing with this paradigm; \Cref{sec:tocuco} presents different important characteristics and statistics of the TOC-UCO datasets, and provide a comparison with the Previous repository; \Cref{sec:5_data_prep} provides the preprocessing and source of each dataset in TOC-UCO; \Cref{sec:6_exp_and_results} presents the first baseline experiments and results on TOC-UCO; Finally, in \Cref{sec:7_conclusions}, the main conclusions and future scope of this work are discussed.}
	
	\rev{
		\section{Ordinal Classification}\label{sec:2_oc}
		In this section, we first formally introduce the OC paradigm, and then specify the main challenges that arise when dealing with this kind of tasks. Note that we include the word \textit{tabular} in the TOC-UCO name to clarify the difference with repositories of other types of data, such as time series \cite{tsoc} or images.
		
		
		\subsection{Problem definition}
		In an OC problem, we are given a set of $K$ independent variables or features $x_i \in \mathcal{X} \subseteq \mathbb{R}^{K}$, with $i = 1, 2, \ldots, N$, ($N$ denotes the number of patterns) associated with a categorical dependent variable $y_i \in \mathcal{Y} = \{\mathcal{C}_1, \mathcal{C}_2, \ldots, \mathcal{C}_Q\}$ (also referred to as class, label, category, or output), which satisfies the ordering constraint $\mathcal{C}_1 \prec \mathcal{C}_2 \prec \ldots \prec \mathcal{C}_Q$, where $\prec$ denotes the order relationship between classes. Here, $Q$ represents the number of classes. The standard strategy in ordinal problems is to map each category in $\mathcal{Y}$ to a numerical value in $\mathbb{N}$ that represents its order. This mapping is commonly defined by a function $v(\mathcal{C}_q) = q$, which
		satisfies $v(\mathcal{C}_1) < v(\mathcal{C}_2) < \ldots < v(\mathcal{C}_Q)$. In this way, an OC dataset can be represented as a set $\mathcal{D} = \{(\mathbf{x}_1, y_1), (\mathbf{x}_2, y_2), \ldots, (\mathbf{x}_i, y_i), \ldots, (\mathbf{x}_N, y_N)\}$.
		
		\rev{Furthermore, two types of OC problems can be distinguished: 1) grouped continuous variables or discretised regression problems, where the output classes are a discrete representation of an original continuous (regression) output. For example, in the \textit{forestfires} dataset \cite{Cortez2007ADM}, the output classes correspond to a discretisation of the area burnt by forest fires, an originally continuous variable expressed in square meters. 2) Assessed ordered categorical variables problems, where no continuous variable are involved, but instead, the output classes are assigned to each pattern by domain experts. For instance, the classes of the \textit{childrenAnemia} dataset represent the level of anaemia suffered by children, diagnosed by doctors. In \Cref{sec:challenges}, the main challenges associated to each type of OC problem are described.}
		
		In \Cref{fig:ordinal_problem}, an intuitive representation of two OC problems is provided. The 2D projection given by the Uniform Manifold Approximation and Projection (UMAP) \cite{mcinnes2018umap} highlights the ordinality of both problems, since the points are distributed spatially according to the arrangement of the classes.
		
		\begin{figure}[hbt!]
			\centering
			\begin{subfigure}{.49\linewidth}
				\includegraphics[width=\linewidth]{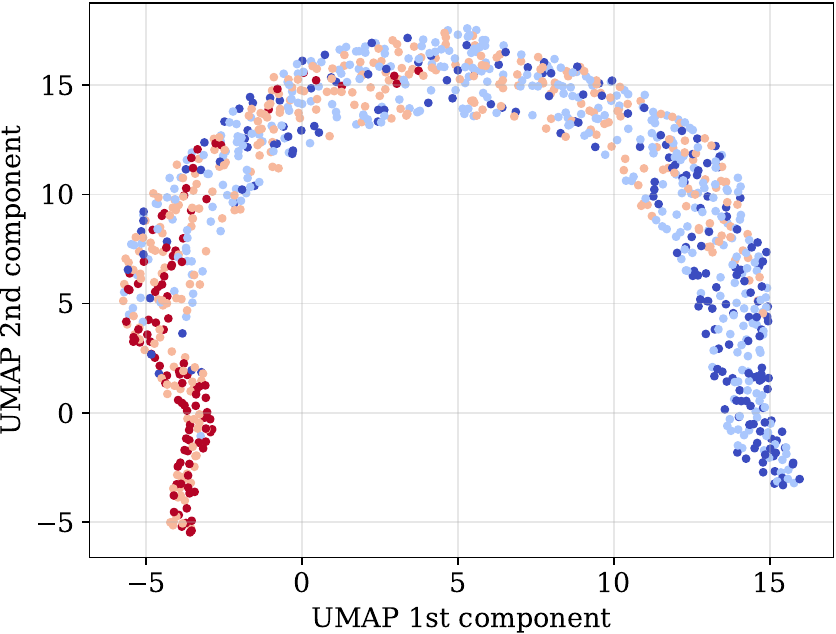}
				\caption{Dataset \textit{oc04\_gymExerciseTracking}.}
			\end{subfigure}\hfill 
			~ 
			\begin{subfigure}{.49\linewidth}
				\includegraphics[width=\linewidth]{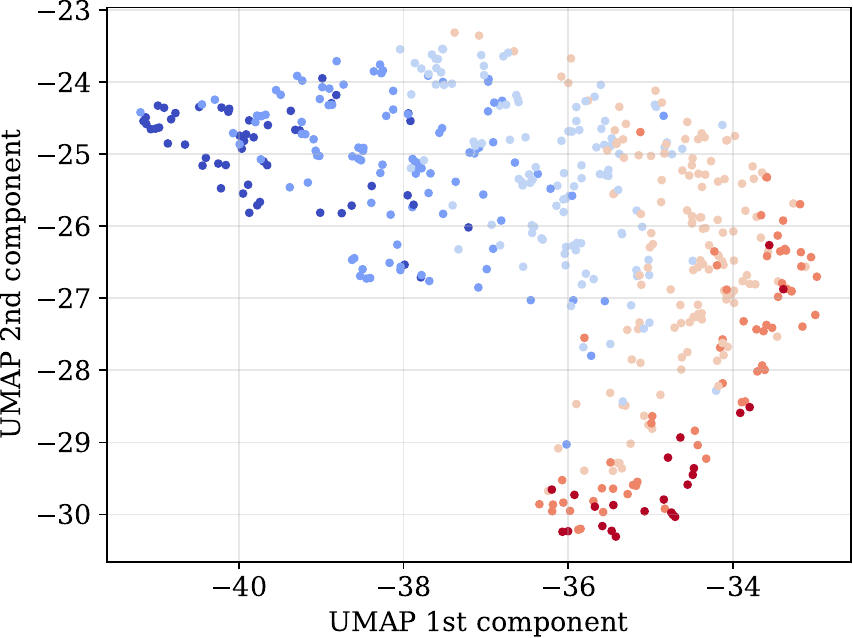}
				\caption{Dataset \textit{oc06\_esl}.}
			\end{subfigure}
			\caption{Uniform Manifold Approximation and Projection (UMAP) of two datasets of TOC-UCO. Points are colored based on the class they belong to, following a gradient from blue (class $\mathcal{C}_1$) to red (class $\mathcal{C}_Q$).}
			\label{fig:ordinal_problem}
		\end{figure}
		
		\subsection{Challenges} \label{sec:challenges}
		Addressing an OC problem involves many important challenges and difficulties, that will also be associated with the specific nature of the task at hand. One of the main challenges comes with the inherent imbalance of ordinal problems, as the extreme classes (e.g. $\mathcal{C}_1$ or $\mathcal{C}_Q$) are naturally less represented, as it is more difficult to collect extreme patterns than regular patterns. For this reason, it is paramount to utilise appropriate metrics that take into account the potential imbalanced performance of the models (e.g. a model can ignore extreme cases and predict only the most represented classes). 
		
		Another important challenge is dealing with the existence of noise in the class assignment. Specifically, when the number of classes is high and the patterns are labelled by several domain experts. For example, in the \textit{winequalityRed} dataset \cite{orreview} included in TOC-UCO, different experts associate bottles of wine with a quality score. Naturally, the subjectivity of this task can lead to noise in the labelling, and thus make it difficult for models to converge. \rev{In this regard, soft labels are one of the more efficient strategies to deal with this kind of problem \cite{vargas2024ebano}. Soft labels replace the one-hot encoding of the target variable by a vector of probabilities where multiple classes can be associated with a non-zero probability. In this way, the noise in the labelling is reflected by the encoding of the target.}
		
		Furthermore, in the specific case of discretised regression tasks, since we are using thresholds in the real line to delimit the ordinal classes, patterns whose original regression output is close to these thresholds are very challenging to learn, given that they belong to a kind of middle-ground between a pair of ordinal classes. For this reason, it is advisable to employ an appropriate binning strategy, such as the k-means discretisation utilised in TOC-UCO.
	}
	
	\section{The TOC-UCO repository}\label{sec:tocuco}
	In this section, we show the main characteristics of TOC-UCO and compare it with the Previous OC repository.
	
	The TOC-UCO repository is publicly available to download\footnote{\url{https://www.uco.es/grupos/ayrna/tocuco}}. Note that a prefix with format ``\{\textit{type}\}\{$Q$\}'' is added to the name of the datasets. Here,  \textit{type} is \textit{oc} for assessed ordered categorical variables problems and \textit{dr} for discretised regression datasets, and $Q$ is the number of classes. This prefix helps to better sort and group the datasets in a computer file system.
	
	\subsection{Characteristics of the TOC-UCO repository}\label{sec:3_tocuco_characteristics}
	
	The characteristics of each dataset in TOC-UCO are presented in \Cref{tab:tocuco}. It can be observed that we have a total of 24 discretised regression datasets and 22 assessed ordered categorical variables problems. Regarding the size of the problems, the smallest dataset is \textit{tae}, with $105$ patterns in the training set and $46$ in the test set, and the largest dataset is \textit{census}, comprising $15948$ patterns in training and $6836$ in test. Note that for the train-test partitions, we followed a stratified 70\%--30\% hold-out. In order to facilitate reproducibility of results, for each dataset, we provide in the repository the indices of 30 different resampled training-test partitions, each computed using a different random seed, numbered from 0 to 29. Instructions for utilising these indices are provided in a tutorial notebook\footnote{\url{https://github.com/ayrna/tocuco/blob/main/tutorial.ipynb}}. Furthermore, in \Cref{tab:comparison}, more statistics on the TOC-UCO datasets, such as mean number of patterns (Mean \#Patterns) or mean imbalance ratio \cite{perez2016ordinalimbalance} are presented.

	\begin{table}
		\footnotesize
		\setlength{\tabcolsep}{2.5pt}
		\renewcommand{\arraystretch}{0.75}
		\caption{Characteristics of the TOC-UCO datasets. $K$ represents the number of input variables. $Q$ denotes the number of classes. The class distribution represents the percentage of patterns that belong to each class, so that the first value corresponds to class 1, second to class 2, etc. The imbalance ratio ($IR$) was computed based on the formula presented in Equation 3 in \cite{perez2016ordinalimbalance}, where values equal to $1 - \frac{1}{Q}$ denote a perfectly balanced problem.}
		\label{tab:tocuco}
		\centering
		\begin{tabular}{lcccccc}
			\toprule
			\multicolumn{7}{c}{Discretised regression datasets}\\
			\midrule
			Dataset                        & \#Train & \#Test & $K$ & $Q$ & Class distribution                                            & $IR$ \\
			\midrule
			forestfires              & 361                & 156               & 8           & 4          & (0.75 0.13 0.04 0.07)                               & 2.63             \\
			machine                  & 146                & 63                & 6           & 4          & (0.70  0.14 0.08 0.08)                               & 1.83             \\
			buoysFlux46026           & 4090               & 1754              & 8           & 5          & (0.39 0.24 0.15 0.11 0.10 )                          & 1.09             \\
			buoysFlux46059           & 4090               & 1754              & 8           & 5          & (0.45 0.25 0.13 0.08 0.09)                          & 1.30             \\
			census1                  & 15948              & 6836              & 8           & 5          & (0.46 0.24 0.13 0.08 0.09)                          & 1.30             \\
			census2                  & 15948              & 6836              & 16          & 5          & (0.46 0.24 0.13 0.08 0.09)                          & 1.30            \\
			buoysFlux46069           & 4090               & 1754              & 8           & 6          & (0.34 0.24 0.16 0.10  0.07 0.09)                     & 1.18             \\
			calhousing               & 14448              & 6192              & 8           & 7          & (0.24 0.19 0.16 0.14 0.10  0.08 0.09)                & 1.01             \\
			buoysHeight46026         & 4090               & 1754              & 8           & 8          & (0.20  0.16 0.16 0.12 0.11 0.09 0.06 0.10 )           & 1.00             \\
			buoysHeight46069         & 4090               & 1754              & 8           & 8          & (0.21 0.18 0.14 0.13 0.11 0.09 0.05 0.09)           & 1.05             \\
			cancerTreatment  & 748                & 321               & 12          & 8          & (0.31 0.14 0.11 0.15 0.09 0.06 0.07 0.07)           & 1.15             \\
			abalone                  & 2923               & 1254              & 10          & 9          & (0.20  0.14 0.16 0.15 0.12 0.06 0.05 0.03 0.09)      & 1.30             \\
			bank1                    & 5734               & 2458              & 8           & 9          & (0.29 0.14 0.12 0.10  0.08 0.07 0.06 0.05 0.09)      & 1.16             \\
			buoysHeight46059         & 4090               & 1754              & 8           & 9          & (0.20  0.16 0.14 0.12 0.10  0.08 0.07 0.05 0.09)      & 1.06             \\
			computer1                & 5734               & 2458              & 12          & 9          & (0.07 0.02 0.03 0.06 0.08 0.11 0.17 0.21 0.25)      & 1.75             \\
			computer2                & 5734               & 2458              & 21          & 9          & (0.07 0.02 0.03 0.06 0.08 0.11 0.17 0.21 0.25)      & 1.75             \\
			housing                  & 354                & 152               & 13          & 9          & (0.18 0.12 0.19 0.17 0.12 0.07 0.05 0.04 0.06)      & 1.21             \\
			insurance                & 936                & 402               & 9           & 9          & (0.28 0.19 0.16 0.12 0.06 0.05 0.03 0.03 0.08)      & 1.55             \\
			soybean                  & 224                & 96                & 9           & 9          & (0.21 0.18 0.13 0.12 0.07 0.07 0.08 0.05 0.08)      & 1.10             \\
			bank2                    & 5734               & 2458              & 32          & 10         & (0.45 0.13 0.09 0.07 0.05 0.04 0.03 0.03 0.03 0.07) & 1.83             \\
			cancerDeathRate          & 2132               & 915               & 29          & 10         & (0.10  0.09 0.10  0.11 0.12 0.13 0.11 0.08 0.06 0.10 ) & 0.95             \\
			concreteStrength         & 721                & 309               & 8           & 10         & (0.14 0.08 0.13 0.14 0.10  0.11 0.08 0.07 0.06 0.09) & 0.97             \\
			realState                & 289                & 125               & 6           & 10         & (0.10  0.09 0.10  0.09 0.11 0.13 0.11 0.09 0.07 0.11) & 0.93             \\
			stock                    & 665                & 285               & 9           & 10         & (0.14 0.08 0.06 0.12 0.11 0.10  0.09 0.08 0.08 0.13) & 0.97             \\
			\midrule
			\multicolumn{7}{c}{Originally ordinal classification datasets}\\
			\midrule
			Dataset                        & \#Train & \#Test & K & Q & Class distribution                                            & IR \\
			\midrule
			balanceScale             & 437                & 188               & 4           & 3          & (0.46 0.08 0.46)                                    & 1.57             \\
			mammoexp                 & 288                & 124               & 5           & 3          & (0.57 0.25 0.18)                                    & 0.92             \\
			newthyroid               & 150                & 65                & 5           & 3          & (0.14 0.7  0.16)                                    & 1.30             \\
			tae                      & 105                & 46                & 54          & 3          & (0.32 0.33 0.34)                                    & 0.67             \\
			car                      & 1209               & 519               & 21          & 4          & (0.7  0.22 0.04 0.04)                               & 3.35             \\
			childrenAnemia           & 4098               & 1757              & 14          & 4          & (0.28 0.28 0.41 0.03)                               & 2.16             \\
			gymExerciseTracking      & 681                & 292               & 17          & 4          & (0.2  0.38 0.31 0.1 )                               & 1.02             \\
			heartDisease             & 205                & 89                & 13          & 4          & (0.64 0.13 0.09 0.15)                               & 1.48             \\
			LESTSensors              & 3578               & 1534              & 6           & 4          & (0.16 0.14 0.23 0.46)                               & 0.98             \\
			LEVXSensors              & 3578               & 1534              & 6           & 4          & (0.33 0.08 0.1  0.48)                               & 1.45             \\
			problematicInternet & 1484               & 636               & 32          & 4          & (0.59 0.27 0.13 0.01)                               & 5.66             \\
			support                  & 511                & 219               & 19          & 4          & (0.39 0.13 0.08 0.4 )                               & 1.34             \\
			swd                      & 700                & 300               & 10          & 4          & (0.03 0.35 0.4  0.22)                               & 2.33             \\
			eucalyptus               & 515                & 221               & 91          & 5          & (0.24 0.15 0.18 0.29 0.14)                          & 0.88             \\
			lev                      & 700                & 300               & 4           & 5          & (0.09 0.28 0.4  0.2  0.03)                          & 2.16             \\
			nhanes                   & 3656               & 1567              & 30          & 5          & (0.11 0.28 0.4  0.18 0.04)                          & 1.72             \\
			vlbw                     & 120                & 52                & 19          & 5          & (0.15 0.17 0.15 0.28 0.25)                          & 0.88             \\
			winequalityRed           & 1119               & 480               & 11          & 5          & (0.04 0.43 0.4  0.12 0.01)                          & 4.88             \\
			esl                      & 341                & 147               & 4           & 6          & (0.11 0.2  0.24 0.28 0.13 0.05)                     & 1.25             \\
			studentPerformance       & 463                & 199               & 43          & 8          & (0.07 0.04 0.11 0.24 0.22 0.16 0.11 0.06)           & 1.31
			\\
			era                      & 700                & 300               & 4           & 9          & (0.09 0.14 0.18 0.17 0.16 0.12 0.09 0.03 0.02)      & 1.66             \\
			melbourneAirbnb          & 14025              & 6011              & 48          & 10         & (0.08 0.09 0.1  0.09 0.05 0.12 0.11 0.09 0.09 0.18) & 1.01            
			\\
			\bottomrule
		\end{tabular}
	\end{table}
	
	\subsection{Comparison with the Previous repository} \label{sec:4_tocuco_vs_prev}
	In this section, we provide a comparison of the proposed TOC-UCO against Previous, the repository introduced in \cite{orreview}, which can be considered the current standard in the OC literature.
	
\begin{figure}[!ht]
		\centering
		\includegraphics[width=\linewidth]{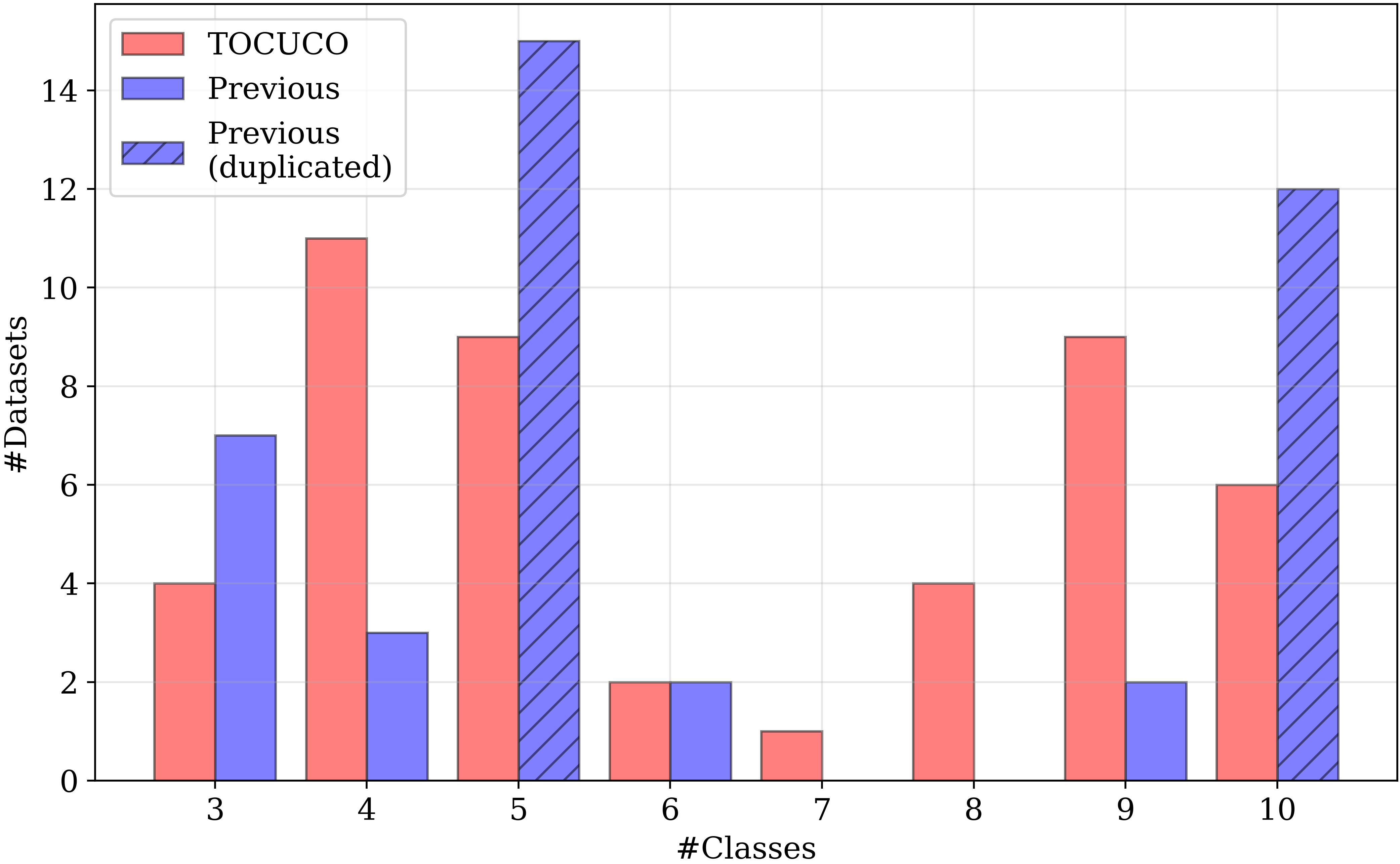}
		\caption{\rev{Comparison between TOC-UCO and Previous repository in terms of the number of datasets per number of classes. Striped bars in the 5 and 10 classes groups of the Previous repository indicate the presence of duplicated datasets (many datasets in Previous are discretised for 5 and 10 classes).}}
		\label{fig:tocuco_vs_previous_n_classes}
	\end{figure}

\begin{figure}[!ht]
		\centering
		\includegraphics[width=\linewidth]{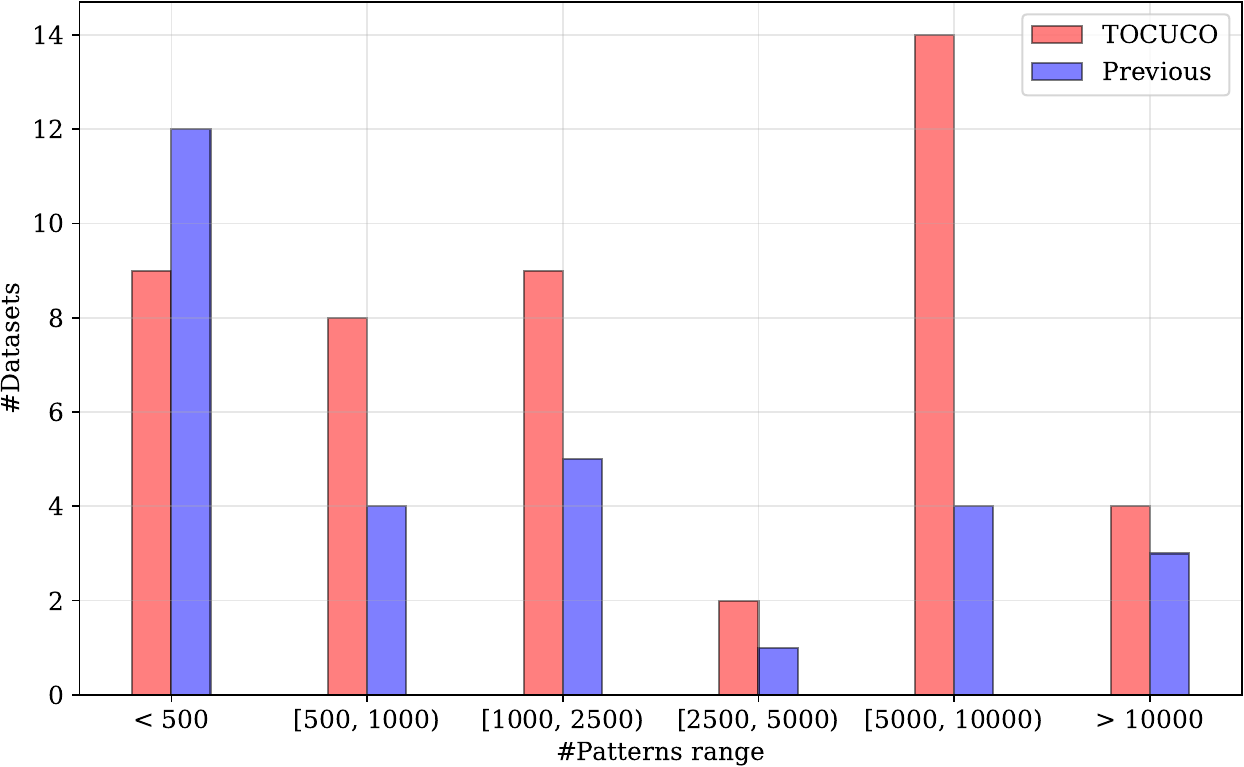}
		\caption{\rev{Comparison between TOC-UCO and Previous repository in terms of the number of datasets per number of patterns. Duplicated datasets in the Previous repository are included only once.}}
		\label{fig:tocuco_vs_previous_n_patterns}
	\end{figure}

	\rev{In \Cref{tab:comparison} we compare the main characteristics of both repositories. It can be observed that, apart from having a larger number of databases, the TOC-UCO archive presents a better distribution of datasets by number of classes, with 16 datasets with a number of classes between 6 and 9, while the Previous repository had only 4. However, given that in the Previous repository each regression dataset was discretised for 5 and 10 classes, it presents a higher number of cases with 10 or more classes. This comparison is graphically represented in \Cref{fig:tocuco_vs_previous_n_classes}, where the striped bars indicate the existence of duplicated datasets in the 5 and 10 classes groups. In the second part of \Cref{tab:comparison}, we observe summary statistics for the number of patterns of both repositories. TOC-UCO presents better mean, minimum, total and number of patterns of the less represented class values. Remarkably, the total number of patterns, obtained by summing up the number of patterns of each dataset, is nearly the double in TOC-UCO. This comparison can be visualised in \Cref{fig:tocuco_vs_previous_n_patterns}, where the datasets are grouped according to different ranges of the number of patterns. It can be observed that TOC-UCO only have less datasets in the ``$<500$'' group, i.e. have a lower number of small datasets. However, TOC-UCO widely surpass the Previous repository in the remaining groups. Concerning the imbalance ratio (IR), Previous presents a higher maximum IR, given that it has one dataset where a class is represented by just one pattern. As mentioned above, this issue is mitigated in TOC-UCO. Regarding the mean IR, TOC-UCO presents a higher value than the Previous archive, this is mainly attributed to the discretisation technique employed: the k-means based discretisation of TOC-UCO \rev{(described in \Cref{sec:discretisation})} returns a more realistic class distribution, which is naturally more imbalanced than the distribution obtained from an equal-frequency discretisation.}
	
	\begin{table}[!ht]
		\centering
		\caption{Comparison between TOC-UCO and the  Previous ordinal classification archive utilised in \cite{orreview}.}
		\label{tab:comparison}
		\begin{tabular}{lcc}
			\toprule
			Feature & Previous & TOC-UCO \\
			\midrule
			\#Unique Datasets & 29 & 46 \\ 
			\#Datasets with 3 classes & 7 & 4 \\
			\#Datasets with 4 classes & 3 & 11 \\
			\#Datasets with 5 classes & 15 & 9 \\
			\#Datasets with 6 to 9 classes & 4 & 16 \\
			\#Datasets with 10 or more classes & 12 & 6 \\
			\midrule
			Mean \#Patterns & 3936 & 4410 \\
			Min \#Patterns & 24 & 151 \\
			Max \#Patterns & 22784 & 22784 \\
			Total \#Patterns & 114160 & 202860 \\
			\#Patterns of the less represented class & 1 & 16 \\
			\midrule
			\makecell[l]{Binning method \\for discretised datasets} & Equal frequency & $k$-means \\
			Duplicated datasets & Yes & No \\
			\midrule
			Mean Imbalance Ratio & 1.401 & 1.552 \\
			Min Imbalance Ratio & 0.667 & 0.667 \\
			Max Imbalance Ratio & 7.939 & 5.660\\
			\bottomrule
		\end{tabular}
	\end{table}
	
	\section{Data source and preprocessing} \label{sec:5_data_prep}
	In this section we describe the source and the preprocessing for each dataset in TOC-UCO. In \Cref{tab:tocuco_sources}, a description of the output ordinal variable and the source of each dataset are presented.
	
	\begin{table}
		\footnotesize
		\setlength{\tabcolsep}{2.5pt}
		\renewcommand{\arraystretch}{0.75}
		\caption{\rev{Brief description of the output of each dataset in TOC-UCO, together with the web and academic sources, grouped by application domain. Sources are provided with an hyperlink, please refer to the online version of the article.}}
		\label{tab:tocuco_sources}
		\centering
		\begin{tabular}{llllc}
			\toprule
			& Dataset & Target description & Source & Reference \\
			\midrule
			\multirow{8}{*}{\rotatebox[origin=c]{90}{Atmospheric}} & buoysFlux46026 & Energy flux & \href{https://www.ndbc.noaa.gov/}{source} & \cite{GOMEZORELLANA2024108462} \\
			& buoysFlux46059 & Energy flux & \href{https://www.ndbc.noaa.gov/}{source} & \cite{GOMEZORELLANA2024108462} \\
			& buoysFlux46069 & Energy flux & \href{https://www.ndbc.noaa.gov/}{source} & \cite{GOMEZORELLANA2024108462} \\
			& buoysHeight46026 & Wave height & \href{https://www.ndbc.noaa.gov/}{source} & \cite{GOMEZORELLANA2024108462} \\
			& buoysHeight46069 & Wave height & \href{https://www.ndbc.noaa.gov/}{source} & \cite{GOMEZORELLANA2024108462} \\
			& buoysHeight46059 & Wave height & \href{https://www.ndbc.noaa.gov/}{source} & \cite{GOMEZORELLANA2024108462} \\
			& LESTSensors & Visibility level & \href{https://ogimet.com/}{source} & --- \\
			& LEVXSensors & Visibility level & \href{https://ogimet.com/}{source} & --- \\
			\midrule
			\multirow{8}{*}{\rotatebox[origin=c]{90}{Price estimation}} & census1 & Median house price & \href{https://www.dcc.fc.up.pt/~ltorgo/Regression/census.html}{source} & \cite{orreview} \\
			& census2 & Median house price & \href{https://www.dcc.fc.up.pt/~ltorgo/Regression/census.html}{source} & \cite{orreview} \\
			& calhousing & Median house value & \href{https://github.com/renatopp/arff-datasets/blob/master/regression/cal.housing.arff}{source} & \cite{orreview} \\
			& realState & House price by unit area & \href{https://archive.ics.uci.edu/dataset/477/real+estate+valuation+data+set}{source} & \cite{orreview} \\
			& stock & Stock price of aerospace companies & \href{https://github.com/renatopp/arff-datasets/blob/master/regression/stock.arff}{source} & \cite{orreview} \\
			& housing & Median owner-occupied home value (in 1000\$s) & \href{https://github.com/renatopp/arff-datasets/blob/master/regression/housing.arff}{source} & \cite{orreview} \\
			& melbourneAirbnb & Price labels for Airbnb rentals & \href{https://huggingface.co/datasets/james-burton/melbourne_airbnb_ordinal}{source} & --- \\
			& insurance & Derived medical costs & \href{https://www.kaggle.com/datasets/mirichoi0218/insurance}{source} & --- \\
			\midrule
			\multirow{10}{*}{\rotatebox[origin=c]{90}{Health}} & cancerTreatmentResponse & Treatment response & \href{https://www.kaggle.com/datasets/lijsbeth/cancer-treatment-response?select=treatment_response_development.csv}{source} & \cite{biomedicines10112679} \\
			& cancerDeathRate & Death rate & \href{https://www.kaggle.com/datasets/varunraskar/cancer-regression/data}{source} & --- \\
			& newthyroid & Thyroid disease stage & \href{https://www.openml.org/search?type=data&status=active&id=40682}{source} & \cite{orreview} \\
			& mammoexp & Mammography experience & --- & \cite{janitza2016random} \\
			& childrenAnemia & Children anemia level & \href{https://www.kaggle.com/datasets/adeolaadesina/factors-affecting-children-anemia-level/data}{source} & --- \\
			& heartDisease & Heart disease stage & \href{https://archive.ics.uci.edu/dataset/45/heart+disease}{source} & \cite{detrano1989international} \\
			& problematicInternetUsage & Severity Impairment Index & \href{https://www.kaggle.com/competitions/child-mind-institute-problematic-internet-use/data}{source} & --- \\
			& support & Functional disability level & --- & \cite{janitza2016random} \\
			& vlbw & Apgar score & --- & \cite{janitza2016random} \\
			& nhanes & Reported general health status & --- & \cite{janitza2016random} \\
			\midrule
			\multirow{7}{*}{\rotatebox[origin=c]{90}{Quality Eval.}} & tae & Teaching performance score & \href{https://github.com/renatopp/arff-datasets/blob/master/classification/tae.arff}{source} & \cite{orreview} \\
			& car & Car quality score & \href{https://github.com/renatopp/arff-datasets/blob/master/classification/car.arff}{source} & \cite{orreview} \\
			& eucalyptus & Seedlots quality & \href{https://github.com/renatopp/arff-datasets/blob/master/agridata/eucalyptus.arff}{source} & \cite{orreview} \\
			& winequalityRed & Red wine quality index & \href{https://www.openml.org/search?type=data&status=active&id=41337}{source} & \cite{orreview} \\
			& studentPerformance & Student final grades & \href{https://archive.ics.uci.edu/dataset/320/student+performance}{source} & \cite{cortez2008using} \\
			& lev & Lecturer performance & \href{https://www.openml.org/search?type=data&status=active&id=1029}{source} & \cite{orreview} \\
			& era & Subjective applicant acceptance Probability & \href{https://www.openml.org/search?type=data&status=active&id=1030&sort=runs}{source} & \cite{orreview} \\
			\midrule
			\multirow{13}{*}{\rotatebox[origin=c]{90}{Other}} & forestfires & Burned area of forest fires & \href{https://archive.ics.uci.edu/dataset/162/forest+fires}{source} & \cite{Cortez2007ADM} \\
			& machine & Published CPU relative performance & \href{https://github.com/renatopp/arff-datasets/blob/master/regression/machine.cpu.arff}{source} & \cite{orreview} \\
			& abalone & Age of abalone & \href{https://github.com/renatopp/arff-datasets/blob/master/regression/abalone.arff}{source} & \cite{orreview} \\
			& bank1 & Fraction of churned customers & \href{https://github.com/renatopp/arff-datasets/blob/master/regression/bank32nh.arff}{source} & \cite{orreview} \\
			& bank2 & Fraction of churned customers & \href{https://github.com/renatopp/arff-datasets/blob/master/regression/bank8FM.arff}{source} & \cite{orreview} \\
			& computer1 & User mode CPU runtime & \href{https://www.dcc.fc.up.pt/~ltorgo/Regression/comp.html}{source} & \cite{orreview} \\
			& computer2 & Restricted attributes usage & \href{https://www.dcc.fc.up.pt/~ltorgo/Regression/comp.html}{source} & \cite{orreview} \\
			& soybean & Number of legumes per plant & \href{https://archive.ics.uci.edu/dataset/913/forty+soybean+cultivars+from+subsequent+harvests}{source} & \cite{de2023dataset} \\
			& concreteStrength & Concrete compressive strength & \href{https://archive.ics.uci.edu/dataset/165/concrete+compressive+strength}{source} & \cite{yeh1998modeling} \\
			& balanceScale & Balance scale tip orientation & \href{https://github.com/renatopp/arff-datasets/blob/master/classification/balance.scale.arff}{source} & \cite{orreview} \\
			& gymExerciseTracking & Workout frequency & \href{https://www.kaggle.com/datasets/valakhorasani/gym-members-exercise-dataset/data}{source} & --- \\
			& swd & Home children risk  & \href{https://www.openml.org/search?type=data&status=active&id=1028}{source} & \cite{orreview} \\
			& esl & Candidate fitness to job & \href{https://www.openml.org/search?type=data&status=active&id=1035}{source} & \cite{orreview} \\
			\bottomrule
		\end{tabular}
	\end{table}
	
	In the following, we describe the preprocessing conducted for each dataset.
	
	\subsection{Discretised regression datasets} \label{sec:discretisation}
	From the set of 24 discretised regression datasets considered in TOC-UCO, a total of 11 datasets were extracted from the Previous OC repository (the one utilised in \cite{orreview}). These datasets are: \textit{machine}, \textit{housing}, \textit{calhousing}, \textit{stock}, \textit{census1} (named \textit{census} in Previous), \textit{census2} (named \textit{census'} in Previous), \textit{bank1} (named \textit{bank} in Previous), \textit{bank2} (named \textit{bank'} in Previous), \textit{abalone}, \textit{computer1} (named \textit{computer} in Previous) and \textit{computer2} (named \textit{computer'} in Previous). \rev{However, we employed a different discretisation (also called binning) strategy of the original regression output variable. In this way, instead of an equal-frequency binning, we applied a k-means based discretisation, so that the values in each bin have the same nearest centre of a 1D k-means clustering \cite{huang2023house} (using the original real output as feature). As a previous step, in order to make this discretisation unaffected by outliers, we filtered the patters whose output is below the 5\% percentile, or above the 95\% percentile. After the discretisation, we reincorporate those outliers and correspondingly assign them to the lowest or highest classes (depending on the percentile they belong to). Concerning the chosen number of classes, we followed the same strategy for each dataset, that consists of selecting the value that returns the lowest Corrected Imbalance Ratio (CIR), that can be expressed as:
		\begin{equation}
			\text{CIR} = \frac{1}{\omega(Q)} \cdot \frac{1}{Q \cdot (Q - 1)}\sum_{q=1}^Q \frac{\sum_{j \neq q} N_j}{N_q},
		\end{equation}
		where $\omega(Q)$ is a correction term that grows linearly with the number of classes. For example, when $Q = 4$, no correction is applied given that $\omega(Q) = 1$, and when $Q = 10$, a correction of $\omega(Q) = 1.25$ is applied to counterbalance the high number of classes. The computation of $\omega(Q)$ is formally expressed as:
		\begin{equation}
			\omega(Q) = \frac{0.25}{6}(Q - 4) + 1
		\end{equation}
	}
	
\begin{figure}[hbt!]
		\centering
		\begin{subfigure}{.49\linewidth}
			\includegraphics[width=\linewidth]{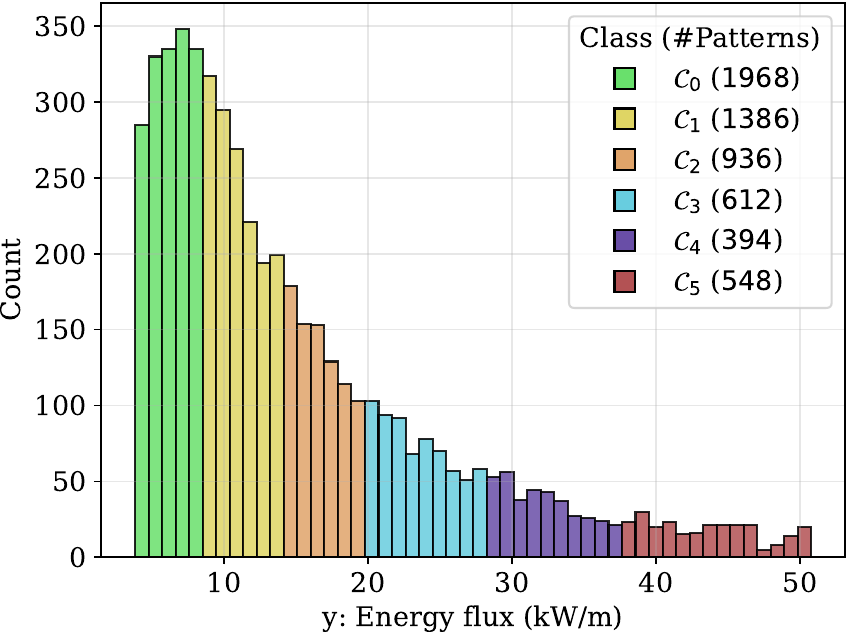}
			\caption{K-means discretised output.}
		\end{subfigure}\hfill 
		~ 
		\begin{subfigure}{.49\linewidth}
			\includegraphics[width=\linewidth]{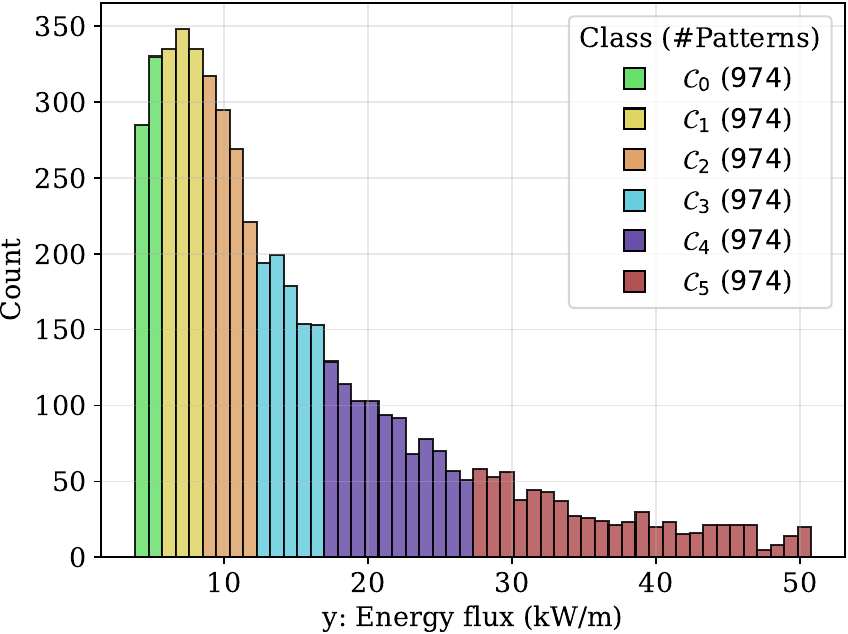}
			\caption{Equal-frequency discretised output.}
		\end{subfigure}
		\caption{\rev{Histogram of the original regression output of the \textit{dr06\_buoysFlux46069} dataset for both discretisation strategies: k-means (utilised in TOC-UCO) and equal-frequency (utilised in  Previous repository). Outliers below 5\% and above 95\% percentiles are discarded to obtain a better histogram shape.}}
		\label{fig:disc_reg_comparison_histogram}
	\end{figure}

	\rev{The k-means based discretisation strategy employed in TOC-UCO results in a more natural distribution of the classes, better suited to ordinal scenarios where the extreme classes are naturally less represented. This discretisation process is the same for all the remaining 13 datasets.
		In \Cref{fig:disc_reg_comparison_histogram}, we can observe the aforementioned difference between k-means and equal-frequency discretisation.}
	
	The source and preprocessing of each dataset are presented below.
	
	\subsubsection{Forestfires.}
	The problem is to predict the burned area of forest fires (variable named \textit{area} in the original source) in the north-east region of Portugal, by using meteorological and other data.
	The dataset was obtained from \footnote{\url{https://archive.ics.uci.edu/dataset/162/forest+fires}}. Spatial and temporal columns are discarded given that tabular classification algorithms are not prepared to to take advantage of them.
	
	\subsubsection{BuoysFlux and BuoysHeight datasets.}
	Energy flux and wave height prediction based on buoys information. The prediction horizon considered is 6 hours. The data is the same utilised in \cite{GOMEZORELLANA2024108462}. The original source is the National Data Buoys Center \footnote{\url{https://www.ndbc.noaa.gov/}}.
	
	\subsubsection{Cancer treatment.}
	Prediction of the pathologically complete response after neoadjuvant chemoradiotherapy for oesophageal cancer \cite{toxopeus2015nomogram} (the target variable is named \textit{Complete\_response\_probability} in the original source). The dataset was collected from \footnote{\url{https://www.kaggle.com/datasets/lijsbeth/cancer-treatment-response?select=treatment_response_development.csv}}. Additionally, the following preprocessing was performed:
	\begin{itemize}
		\item Column \textit{Tumor\_type} was converted to a binary column named \textit{adenocarcinoma}, as it has only two values: \textit{adenocarcinoma} and \textit{squamous cell carcinoma}.
		\item The \textit{M\_stage} column was discarded given that it is constant.
		\item The \textit{Smoking} column was also discarded due to the high number of null values.
		\item \textit{Tumor\_location} was converted into multiple dummy columns.
		\item Every row with null values in all columns was removed.
	\end{itemize}
	
	\subsubsection{Insurance.}
	The problem is to predict individual medical costs billed by health insurance (column named \textit{charges} in the original source). The dataset was collected from \footnote{\url{https://www.kaggle.com/datasets/mirichoi0218/insurance}}. The only preprocessing performed was to convert the column \textit{region} into multiple dummy columns and to remove rows with null values, no missing value imputation is performed.
	
	\subsubsection{Soybean.}
	Prediction of the number of legumes per plant in soybean cultivars. The dataset was obtained from \footnote{\url{https://archive.ics.uci.edu/dataset/913/forty+soybean+cultivars+from+subsequent+harvests}}. The preprocessing conducted was to discard column \textit{Cultivar} (as it has too many text unique values), and to remove rows with null values. No missing value imputation is performed.
	
	\subsubsection{CancerDeathRate.}
	Prediction of the cancer death rate in specific regions (column named \textit{target\_deathrate} in the original source). The dataset was obtained from \footnote{\url{https://www.kaggle.com/datasets/varunraskar/cancer-regression/data}}. The following preprocessing was performed:
	\begin{itemize}
		\item Column \textit{binnedic}, which was originally filled with intervals, e.g.: ``$[1232.9, 46432.8)$'', was converted into two columns containing the min and max of the interval, e.g. $\textit{binnedic\_min} = 1232.9$ and $\textit{binnedic\_max} = 46432.8$.
		\item \textit{Geography} column was discarded given that it has almost 3000 unique values.
		\item Columns \textit{pctsomecol18\_24}, \textit{pctemployed16\_over} and \textit{pctprivatecoveragealone} were discarded due to the high number of null values.
	\end{itemize}
	
	\subsubsection{ConcreteStrength.}
	The problem is to predict the compressive strength of concrete (variable named \textit{Concrete compressive strength(MPa, megapascals)} in the original source). The dataset was obtained from \footnote{\url{https://archive.ics.uci.edu/dataset/165/concrete+compressive+strength}}.
	
	\subsubsection{RealState.}
	Prediction of house prices by unit area \cite{realstate}. The dataset was collected from \footnote{\url{https://archive.ics.uci.edu/dataset/477/real+estate+valuation+data+set}}.
	
	\subsection{Originally ordinal classification datasets}
	From the 22 originally ordinal classification datasets in TOC-UCO, a total of 10 datasets were extracted from the repository utilised in \cite{orreview}. These datasets are: \textit{balanceScale}, \textit{newthyroid}, \textit{tae}, \textit{car}, \textit{swd}, \textit{eucalyptus}, \textit{lev}, \textit{winequalityRed}, \textit{esl}, \textit{era}. 
	In the case of \textit{winequalityRed} and \textit{esl}, we alleviated the extreme class imbalance by merging extreme classes. In this way, for \textit{winequalityRed} we merged class 0 with class 1, and for \textit{esl} we merged class 0 and 1 with class 2, and class 8 with 7. In the following, we present the remaining new 12 datasets.
	
	\subsubsection{Mammoexp.}
	The objective is to model the relationship between mammography experience: have never had a mammography (class 0), have had one within the last year (class 1), last mammography greater than one year ago (class 2), and the attitude towards mammography. The dataset was sourced from \cite{janitza2016random}.
	
	\subsubsection{ChildrenAnemia.}
	Prediction of children anemia level based on different physical and socioeconomic factors. The datasets was obtained from \footnote{\url{https://www.kaggle.com/datasets/adeolaadesina/factors-affecting-children-anemia-level/data}}. The following preprocessing was performed:
	\begin{itemize}
		\item Variable \textit{Type of place of residence} is replaced by \textit{residence\_urban\_or\_rural}, a binary variable being: 1 if residence is urban and 0 if rural.
		\item Variable \textit{Highest educational level} is replaced by \textit{education\_level}, an ordinal variable encoding original values as follows: \textit{No education}: 0, \textit{Primary}: 1, \textit{Secondary}: 2, \textit{Higher}: 3.
		\item Variable \textit{Wealth index combined} is replaced by \textit{wealth\_index}, an ordinal variable encoding original values as follows: 
		\textit{Poorest}: 0, \textit{Poorer}: 1, \textit{Middle}: 2, \textit{Richer}: 3, \textit{Richest}: 4.
		\item Variable \textit{Have mosquito} bed net for sleeping (from household questionnaire) is replaced by \textit{mosquito bed net}, a binary variable being: 1 if mosquito bed net is used and 0 if not.
		\item Variable \textit{Smokes cigarettes} is replaced by \textit{smokes\_cigarettes}, a binary variable being: 1 when smoking and 0 if not.
		\item Variable \textit{Current marital status} is replaced by \textit{married}, a binary variable being: 1 if married and 0 if not.
		\item Variable \textit{Currently residing with husband/partner} is replaced by \textit{living\_with\_partner}, a binary variable being: 1 if living with husband or partner and 0 if not.
		\item Variable \textit{Had fever in last two weeks} is replaced by \textit{had\_fever\_in\_last\_two\_weeks}, a binary variable being: 1 if fever in the last two weeks and 0 if not.
		\item Variable \textit{Taking iron pills, sprinkles or syrup} is replaced by \textit{taking\_iron\_pills\_sprinkles\_or\_syrup}, a binary variable being: 1 if taking those pills and 0 if not.
		\item Variable \textit{Age in 5-year groups} is replaced by \textit{age\_group}, an ordinal variable doing the next replacement from original column:
		15-19: 0, 20-24: 1, 25-29: 2, 30-34: 3, 35-39: 4, 40-44: 5, 45-49: 6.
	\end{itemize}
	
	\subsubsection{GymExerciseTracking.}
	The problem is to predict the number of workout sessions per week for gym members (variable named \textit{Workout\_Frequency (days/week)} in the source). The dataset was obtained from \footnote{\url{https://www.kaggle.com/datasets/valakhorasani/gym-members-exercise-dataset/data}}. The preprocessing consisted of transforming the variable \textit{workout\_type} into multiple dummy variables and removing missing values.
	
	\subsubsection{HeartDisease.}
	The objective is to predict the presence of heart disease in patients (variable named \textit{num} at source), ranging from 0 (no heart disease presence) to 4 (severe heart disease). This data is sourced from \footnote{\url{https://archive.ics.uci.edu/dataset/45/heart+disease}}. The preprocessing consisted of removing missing values.
	
	\subsubsection{ProblematicInternetUsage.}
	The problem is to predict the Severity Impairment Index (sii), a standard measure of problematic internet use, based on different demographic and behavioural factors. This data was obtained from \footnote{\url{https://www.kaggle.com/competitions/child-mind-institute-problematic-internet-use/data}}. The preprocessing consisted of:
	\begin{itemize}
		\item Columns with more than a 35\% of missing values were discarded.
		\item Seasonal columns are discarded due to a high number of unique values.
		\item Target variable \textit{sii} is transformed to ordinal by mapping: \textit{None} to class 0, \textit{Mild} to class 1, \textit{Moderate} to class 2, and \textit{Severe} to class 3.
	\end{itemize}
	
	\subsubsection{Support.}
	Prediction of functional disability, which is categorised into 5 ordered classes from slight to severe. This dataset was sourced from \cite{janitza2016random}.
	
	\subsubsection{vlbw.}
	The problem is to predict the Apgar score, a score for the physical health status of a newborn measured on a 9-point scale, from diverse factors such as the weight and sex of the newborn and the type of delivery. The dataset was sourced from \cite{janitza2016random}.
	
	\subsubsection{Nhannes.}
	Prediction of the self-reported general health status from demographical and health-related factors. The response is categorised into five classes (0: \textit{excellent}, 1: \textit{very good}, 2: \textit{good}, 3: \textit{fair}, 4: \textit{poor}). Sourced from \cite{janitza2016random}.
	
	
	\subsubsection{StudentPerformance.}
	Prediction of student grades in Mathematics and Portuguese language subjects. The input features include student grades in other different subjects, demographic, and other social and school related characteristics. Data was collected by using school reports and questionnaires. The dataset was sourced from \footnote{\url{https://archive.ics.uci.edu/dataset/320/student+performance}}. The following preprocessing was performed:
	\begin{itemize}
		\item All nominal variables \textit{Mjob}, \textit{Fjob}, \textit{reason} and \textit{guardian} have been converted into multiple binary (dummy) variables.
		\item Columns \textit{G1} and \textit{G2}, representing the first and second period grades, are discarded given that we want to focus on predicting the final (global) grade.
		\item We reduced the number of classes by grouping the original target values so that: original 0, 1 and 2 values are mapped to 0, 3 and 4 to 1, 5 and 6 to 2, \ldots, 19 and 20 to 10. After this step, some classes remained extremely under-represented. To fix this, we merged class 2 with 1 and class 8 with 9, so that we were left with 8 classes.
	\end{itemize}
	
	\subsubsection{MelbourneAirbnb.}
	The problem is to predict price labels for Airbnb rentals (variable \textit{price\_label} at source). The dataset was obtained from \footnote{\url{https://huggingface.co/datasets/james-burton/melbourne_airbnb_ordinal}}. The preprocessing conducted consisted of:
	\begin{itemize}
		\item All text mining columns were discarded.
		\item Constant valued columns were discarded.
		\item Converted \textit{calendar\_updated} text valued variable to integer variable representing the number of days since the last calendar update. Rows with no update are discarded.
		\item Columns with more than 20\% missing values are discarded. Then, any row with at least one missing value is removed. No missing value imputation is performed.
	\end{itemize}
	
	
		
		
		
		
		
	
	\section{Baseline experiments and results} \label{sec:6_exp_and_results}
	In order to provide the OC literature with a baseline experimentation, we provide the results obtained in the TOC-UCO repository by 8 different methodologies: 
	\begin{itemize}
		\item Ridge. This method converts the classification problem into a multi-output regression task by mapping the target values into $Q$ outputs taking values in the set $\{-1, 1\}$, there the value will be $1$ is the class is the one observed in the pattern \cite{pedregosa2011scikit}.
		\item Random Forest (RF). This method trains an ensemble of decision trees and performs the final prediction by majority voting \cite{breiman2001random}.
		\item EXtreme Gradient Boosting classifier (XGB). XGB is an efficient and scalable implementation of a gradient boosting machine, presented in \cite{xgboost}.
		\item Logistic All-Threshold classifier (LogAT). An OC technique based on the cumulative link models (CLM) \cite{tsoc}, employing the all-thresholds loss presented in \cite{pedregosa2017consistency}.
		\item Multi-Layer Perceptron (MLP). A single-layer neural network trained with the cross-entropy loss \cite{bishop2006mlp}.
		\item MLP-CLM. An MLP model that replaces the softmax by a CLM output layer \cite{vargas2020cumulative,berchez2025dlordinal}.
		\item Triangular MLP (MLP-T). A MLP trained with soft labels computed from a triangular distribution \cite{vargas2023soft,berchez2025dlordinal}.
		\item Ensemble BAsed on uNimodal Ordinal (EBANO) classifier. An emsemble of LogAT, MLP-CLM and MLP-T, where the weights of the ensemble are cross-validated in terms of the Averaged Mean Absolute Error (AMAE) \cite{vargas2024ebano}.
	\end{itemize}
	
	The hyperparameters of the eight methodologies presented above are cross-validated following a 3-fold strategy in the training set. In \Cref{tab:crossvalidation}, the different values considered in the cross-validation of each hyperparameter are specified. Due to the inherent imbalance of ordinal problems commented in \Cref{sec:challenges}, all the methodologies are trained employing balanced class weights, i.e. each sample belonging to a class $\mathcal{C}_q$ is weighted with $\omega_q = \frac{N}{Q \cdot N_q}$, where $N_q$ is the total number of samples of class $\mathcal{C}_q$. The code implementation of these experiments is publicly available in GitHub\footnote{\url{https://github.com/ayrna/tocuco/blob/main/baseline_experiments.py}}.\\
	
	\begin{table}[]
		\centering
		\footnotesize
		\setlength{\tabcolsep}{2.5pt}
		\renewcommand{\arraystretch}{0.75}
		\caption{Hyperparameter values used for cross-validation of for each technique.}
		\label{tab:crossvalidation}
		\begin{tabular}{lll}
			\toprule
			Method            & Hyperparameter           & Values                                                                                       \\
			\midrule
			\multirow{2}{*}{Ridge and LogAT}   & Regularisation strength  & \{$10^{-3}, 10^{-2}, 10^{-1}, 10^0, 10^1, 10^2, 10^3$\}        \\
			& Maximum iterations       & \{$1000, 1500, 3000, 5000$\} \\
			\midrule
			\multirow{5}{*}{RF}                & Max depth                               & \{$3, 5, 8$\},                  \\
			& \#Estimators                            & \{$100, 250, 500, 1000$\}            \\
			& Minimal Cost-Complexity Pruning         & \{$0.0, 0.05, 0.1$\}         \\
			& Feature subsample                       & \{No, squared root\}           \\
			& Bootstrap                               & \{No, Yes\}            \\
			\midrule
			\multirow{5}{*}{XGB}               & Max depth                & \{$3, 5, 8$\}                      \\
			& \#Estimators             & \{$100$, $250$, $500$, $1000$\}      \\
			& Learning rate            & \{$0.01$, $0.05$, $0.10$\} \\
			& Pattern subsample (\%)   & \{$0.75$, $0.95$, $1.0$\}      \\
			& Feature subsample (\%)    & \{$0.75$, $0.95$, $1.0$\}     \\
			\midrule
			MLP,              & \#Hidden units      & \{$5, 8, 10, 15, 20, 50, 100$\} \\
			MLP-CLM and       & \#Epochs            & \{$1000, 1500, 3000, 5000$\} \\
			MLP-T             & Learning rate       & \{$10^{-5}, 10^{-4}, 10^{-3}$\} \\
			\midrule
			MLP-CLM           & Minimum distance          & \{$0.0, 0.1, 0.2$\}  \\
			\midrule
			MLP-T             & Alpha                     & \{$0.05, 0.10$\}    \\
			\bottomrule
		\end{tabular}
	\end{table}
	
\begin{figure}[hbt!]
		\centering
		\begin{subfigure}{.49\linewidth}
			\includegraphics[width=\linewidth]{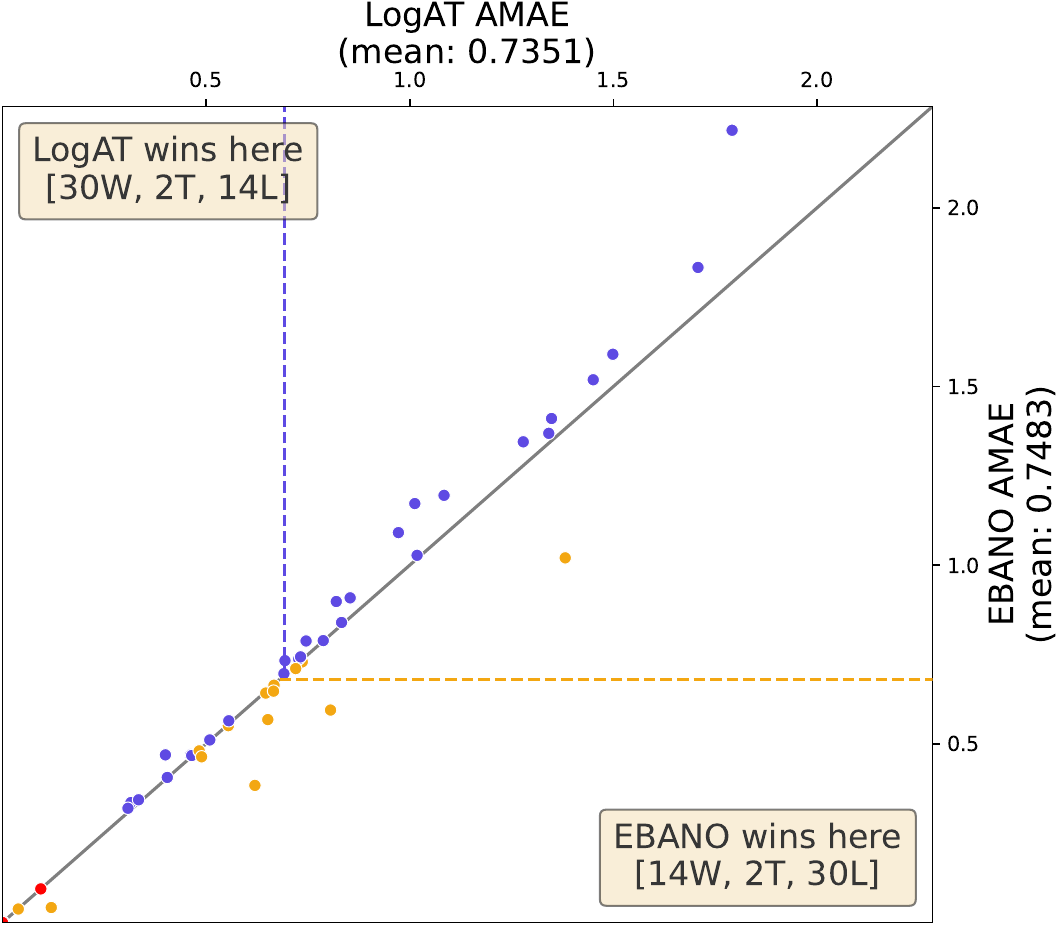}
			\caption{AMAE}
		\end{subfigure}\hfill 
		~ 
		\begin{subfigure}{.49\linewidth}
			\includegraphics[width=\linewidth]{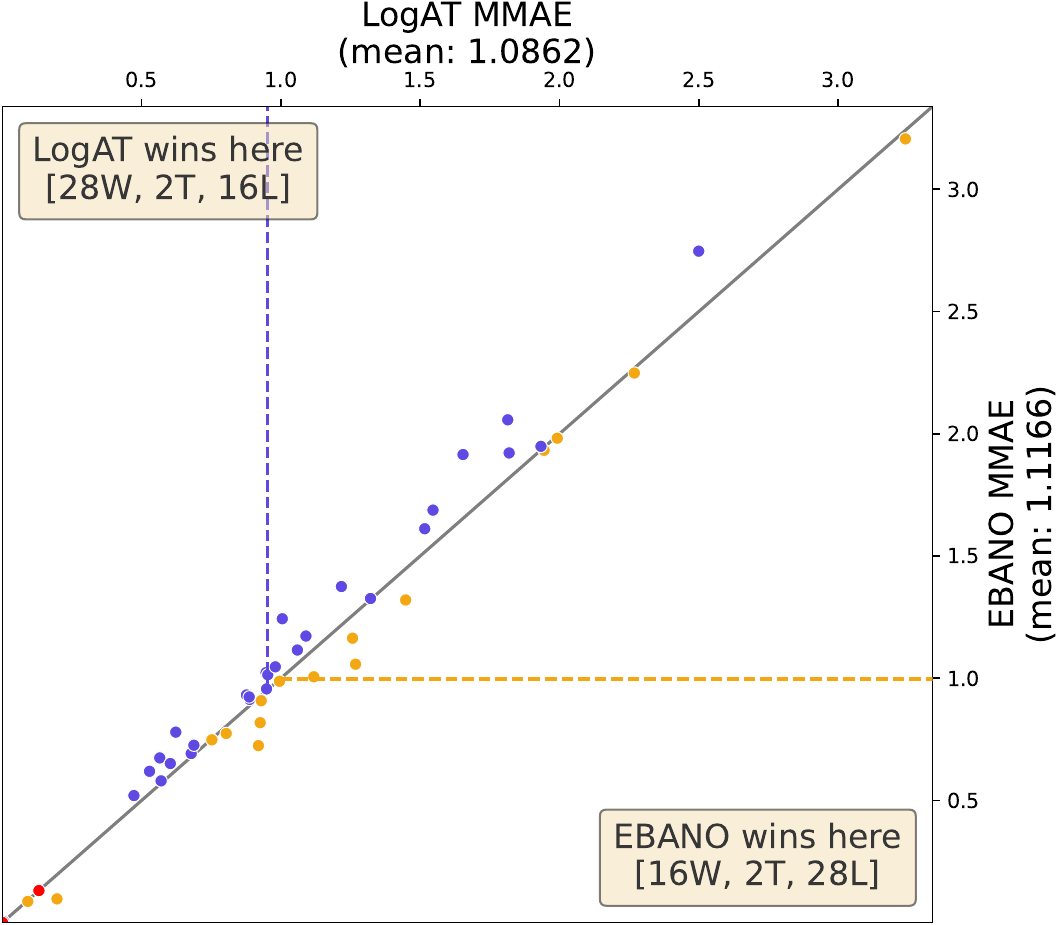}
			\caption{MMAE}
		\end{subfigure}
		
		\medskip 
		\begin{subfigure}{.49\linewidth}
			\includegraphics[width=\linewidth]{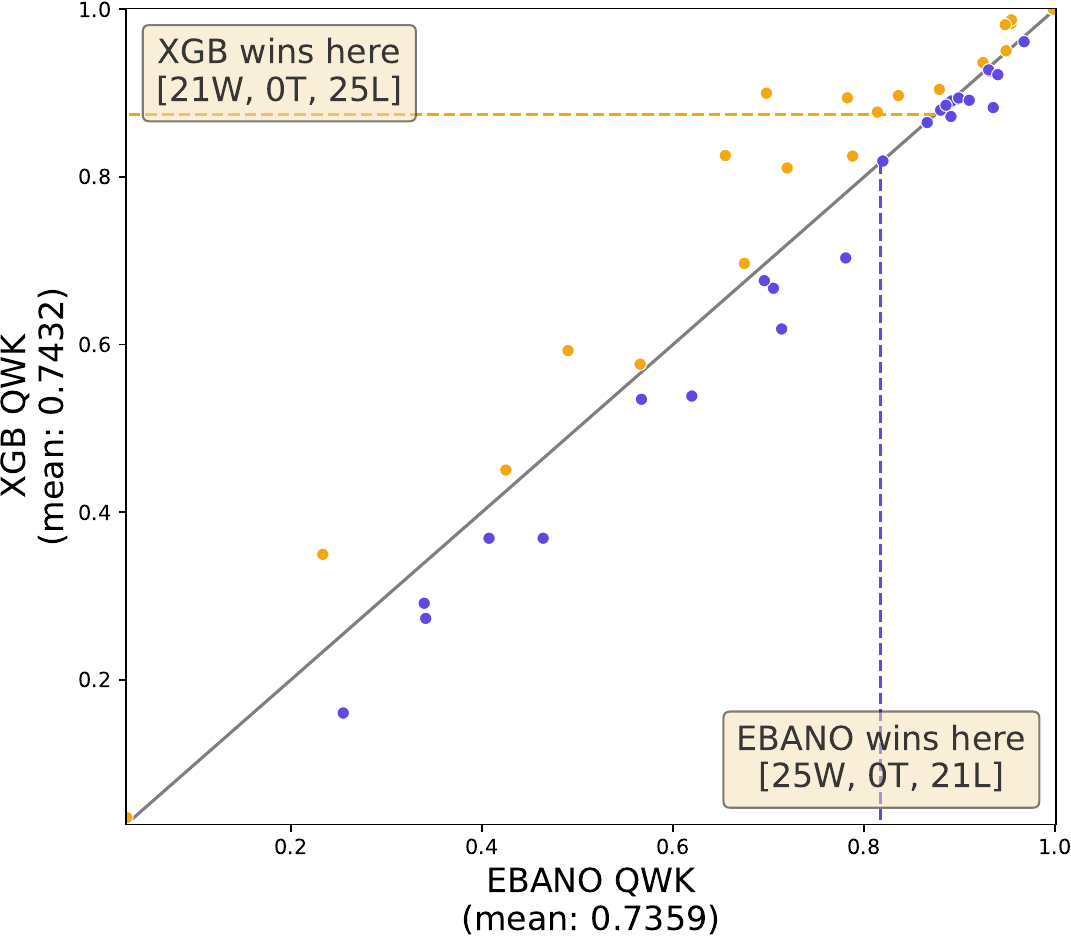}
			\caption{QWK}
		\end{subfigure}\hfill 
		\begin{subfigure}{.49\linewidth}
			\includegraphics[width=\linewidth]{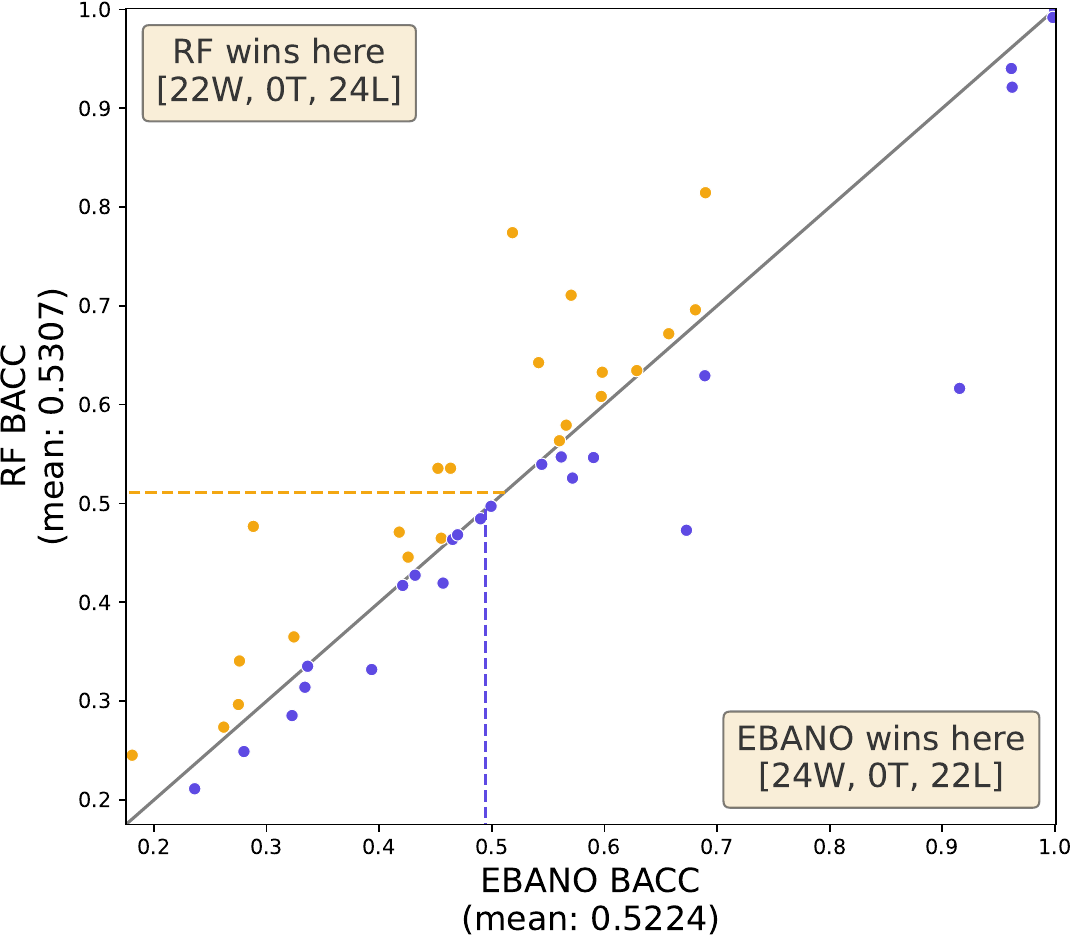}
			\caption{BACC}
		\end{subfigure}
		
		\caption{Scatter plot comparing the two best methods: LogAT and XGB, for each metric. Each point represent the performance obtained in a single dataset of TOC-UCO, positioned so that the horizontal axis coordinate is the value obtained by the LogAT, and the vertical axis coordinate takes the value obtained by XGB. The points drawn above the diagonal line indicate a better performance of the method displayed in that side of the figure. Dashed lines represent the median value for each corresponding method-metric pair. Additionally, in the top-right and bottom-left corners, the number of wins (W), ties (T) and losses (L) are correspondingly displayed for each method. }
		\label{fig:scatter}
	\end{figure}

	The methodologies are evaluated in terms of 4 different performance metrics:
	\begin{itemize}
		\item Averaged Mean Absolute Error (AMAE). This metric accounts for the imbalance of the problem by averaging the isolated MAE obtained for each class \cite{Baccianella2009AMAE}, where the MAE corresponds to the mean absolute distance between true and predicted labels coded as integers (i.e. it corresponds to the mean absolute deviation in number of categories of the ordinal scale).
		\item Maximum Mean Absolute Error (MMAE). This metric follows the same logic as the AMAE, but focuses on the maximum MAE along the classes (worst classified class), instead of the computing the average \cite{CRUZRAMIREZ201421}.
		\item Quadratic Weighted Kappa (QWK), a metric based on the kappa index that measures the inter-rater agreement on classifying elements into a set of classes. Its value ranges from $-1$ to $1$, with $1$ representing a perfect prediction. This metric is weighted with quadratic penalisation values that increase with the distance between classes \cite{vargas2023exponential}.
		\item Balanced Accuracy (BACC), this metric computes the average of the sensitivity (true positive rate) across all classes. It differs from regular accuracy by considering the imbalance in the dataset, giving each class an equal weight in the computation regardless of its size.
	\end{itemize}
	AMAE, MMAE and QWK are OC metrics, while BACC is nominal. The results obtained from the experiments are shown in \Cref{tab:results_mean_std} and \Cref{tab:results_rankings}. In \Cref{tab:results_mean_std}, the results are presented in terms of the mean and standard deviation for each model-metric pair. It can be observed that the RF method obtains the lowest mean in terms of AMAE. However, LogAT is very close with respect to the mean, and presents almost half of the standard deviation obtained by RF, i.e. it is much more robust. With respect to MMAE, LogAT is the best methodology with respect to both mean and standard deviation. EBANO obtains the second best values. In terms of QWK, XGB achieves the highest (best) values, and EBANO the second best results. On the other hand, RF obtains the highest BACC, followed by the XGB. In \Cref{tab:results_rankings}, the rankings are provided. In this table, the LogAT manages to surpass the rest of the methods in terms of AMAE and MMAE. However, with respect to QWK and BACC, EBANO stands as the best performer, followed by XGB in terms of QWK, and by RF in terms of BACC. In \Cref{fig:scatter}, a more detailed comparison of the two best methods for each metric is displayed using the scatter plot provided in the aeon library \cite{middlehurst2024aeon}.
	
	Code and instructions on how to use TOC-UCO and reproduce the experiments are provided in online\footnote{\url{https://github.com/ayrna/tocuco}}.
	
	\begin{table}[h!]
		\setlength{\tabcolsep}{3pt}
		\renewcommand{\arraystretch}{1.35}
		\caption{Mean and standard deviation (Mean$_\text{STD}$) results for the different methodologies and performance metrics. The metrics marked with ($\uparrow$) are to be maximised, while the metrics marked with ($\downarrow$) are to minimise. The best results are highlighted in \textbf{bold}, while the second best are highlighted in \textit{italics}}
		\centering
		\begin{tabular}{lccccc}
			\toprule
			Method      & AMAE ($\downarrow$)   & MMAE ($\downarrow$)   & QWK ($\uparrow$)  & BACC ($\uparrow$) \\
			\midrule
			Ridge       & $1.059_{0.055}$             & $1.765_{0.121}$             & $0.645_{0.028}$           & $0.424_{0.025}$           \\
			MLP-CLM     & $0.960_{0.191}$             & $1.540_{0.411}$             & $0.681_{0.076}$           & $0.469_{0.054}$           \\
			MLP-T       & $0.942_{0.160}$             & $1.553_{0.359}$             & $0.659_{0.079}$           & $0.463_{0.056}$           \\
			MLP         & $0.908_{0.148}$             & $1.467_{0.342}$             & $0.677_{0.070}$           & $0.477_{0.051}$           \\
			XGB         & $0.761_{0.046}$             & $1.295_{0.138}$             & $\mathbf{0.743_{0.025}}$  & $\mathit{0.523_{0.023}}$   \\
			EBANO       & $0.748_{0.051}$             & $\mathit{1.117_{0.117}}$    & $\mathit{0.736_{0.023}}$  & $0.522_{0.029}$           \\
			LogAT       & $\mathit{0.735_{0.035}}$    & $\mathbf{1.086_{0.098}}$    & $0.717_{0.020}$           & $0.479_{0.025}$           \\
			RF          & $\mathbf{0.734_{0.066}}$    & $1.150_{0.167}$             & $0.729_{0.036}$           & $\mathbf{0.531_{0.031}}$   \\
			\bottomrule
		\end{tabular}
		\label{tab:results_mean_std}
	\end{table}
	
	\begin{table}[h!]
		\setlength{\tabcolsep}{3pt}
		\renewcommand{\arraystretch}{1.35}
		\caption{Mean rankings obtained by each methodology in each metric. The best results are highlighted in \textbf{bold}, while the second best are highlighted in \textit{italics}}
		\centering
		\begin{tabular}{lccccc}
			\toprule
			Method          & AMAE	             & MMAE	                & QWK	            & BACC              \\
			\midrule
			LogAT           & $\mathbf{2.620}$   & $\mathbf{2.402}$	    & $3.728$ 	        & $5.033$               \\
			EBANO           & $\mathit{2.772}$   & $\mathit{2.576}$	    & $\mathbf{2.359}$ 	& $\mathbf{2.576}$      \\
			RF              & $3.109$            & $3.326$	            & $3.435$ 	        & $\mathit{2.870}$      \\
			XGB             & $3.935$            & $4.522$	            & $\mathit{2.891}$ 	& $3.478$               \\
			MLP             & $4.978$            & $5.033$	            & $5.565$ 	        & $4.804$               \\
			MLP-CLM         & $5.826$            & $5.587$	            & $5.065$ 	        & $5.522$               \\
			MLP-T           & $5.978$            & $6.054$	            & $6.478$ 	        & $5.957$               \\
			Ridge           & $6.783$            & $6.500$	            & $6.478$ 	        & $5.761$               \\
			\bottomrule
		\end{tabular}
		\label{tab:results_rankings}
	\end{table}

	
	\section{Conclusions} \label{sec:7_conclusions}
	In this work we presented TOC-UCO, a new OC benchmarking repository. TOC-UCO comes to address the limitations of the current literature in OC, where the most common practice adopted when validating new techniques is to consider a subset of the problems utilised in \cite{orreview}, a survey published in 2016. In addition, these datasets utilised in the current literature present many drawbacks, such as the very low number of patterns in some problems, and the discretisation strategy employed to obtain the discretised regression datasets. The TOC-UCO repository comes to address this gap, improving on the Previous repository, and providing an updated set of ordinal problems. The main areas of improvement brought by the TOC-UCO repository are: an increased number of datasets with a sufficiently high number of patterns, a more appropriate discretisation technique for regression datasets, and a wider spectrum with respect to the number of classes.
	
	In order to ensure data transparency, the source and preprocessing of each dataset are presented in this work. Furthermore, to ease the execution and reproduction of experiments with TOC-UCO, we provided 30 different stratified train-test partitions, employing a 70\%-30\% hold-out. Instructions on loading and using these partitions are provided in a GitHub repository\footnote{\url{https://github.com/ayrna/tocuco}}.
	
	In addition, a baseline experimentation in TOC-UCO was also provided in this work, considering 4 widely utilised nominal approaches: Ridge, RF, XGB, and MLP, and 4 state-of-the-art ordinal techniques: LogAT, MLP-CLM, MLP-T, and EBANO. With this purpose, each method was trained and evaluated on each of the train-test partitions mentioned above. The results were evaluated in terms of BACC and 3 ordinal metrics: AMAE, MMAE, and QWK.
	
	This work contributes to the OC literature with an improved and renewed set of benchmark problems, which will ease the validation of new techniques and the general development of this area. We appreciate any contribution to this repository.
	
	\section*{Acknowledgements}
	The present study is supported by the ``Agencia Estatal de Investigación (España)'' (grant ref.: PID2023-150663NB-C22 / AEI / 10.13039 / 501100011033), by the EU Commission, AgriFoodTEF (grant ref.: DIGITAL-2022-CLOUD-AI-02, 101100622), by the Secretary of State for Digitalization and AI ENIA International Chair (grant ref.: TSI-100921-2023-3), and by the University of Córdoba (grant ref.: PP2F\_L1\_15). Rafael Ayllón-Gavilán is supported by the ``Instituto de Salud Carlos III'' (ISCIII) and EU (grant ref.: FI23/00163). Antonio Manuel Gómez-Orellana is supported by ``Consejería de Transformación Económica, Industria, Conocimiento y Universidades de la Junta de Andalucía'' (grant ref.: PREDOC-00489). F. Bérchez-Moreno has been supported by ``Plan Propio de Investigación Submodalidad 2.2 Contratos predoctorales'' of the University of Córdoba.
	
	\bibliographystyle{splncs04}
	\bibliography{bibliography}
	
\end{document}